\documentclass[runningheads]{llncs}

\usepackage{eccv}

\usepackage{eccvabbrv}

\usepackage{graphicx}
\usepackage{booktabs}

\usepackage{makecell}
\usepackage{booktabs}
\usepackage{multirow}
\usepackage{tikz}
\usepackage{wrapfig}
\usepackage{subcaption}
\usepackage{enumitem}

\usepackage{soul}
\usepackage{xcolor}
\sethlcolor{white}

\usepackage[accsupp]{axessibility}  

\usepackage[breaklinks,colorlinks,citecolor=eccvblue]{hyperref}

\usepackage{orcidlink}

\begin{document}

\title{UHD-IQA Benchmark Database: Pushing the Boundaries of Blind Photo Quality Assessment} 

\titlerunning{UHD-IQA Benchmark Database}

\author{Vlad Hosu\inst{1}\orcidlink{0000-0001-7070-5688} \and
Lorenzo Agnolucci\inst{1, 3}\orcidlink{0000-0002-9558-1287} \and
Oliver Wiedemann\inst{2}\orcidlink{0000-0001-9534-126X} \and
Daisuke Iso\inst{1} 
\and Dietmar Saupe\inst{2}\orcidlink{0000-0001-6735-5103}}

\authorrunning{V.~Hosu et al.}

\institute{Sony AI -- \email{[name.surname@sony.com]} \and
University of Konstanz, Germany -- \email{[name.surname@uni-konstanz.de]} \and University of Florence, Italy -- \email{[name.surname@unifi.it]}}

\maketitle

\begin{abstract}
We introduce a novel Image Quality Assessment (IQA) data\-set comprising 6073 UHD-1 (4K) images, annotated at a fixed width of 3840 pixels. Contrary to existing No-Reference (NR) IQA datasets, ours focuses on highly aesthetic photos of high technical quality, filling a gap in the literature. The images, carefully curated to exclude synthetic content, are sufficiently diverse to train general NR-IQA models. Importantly, the dataset is annotated with perceptual quality ratings obtained through a crowdsourcing study. Ten expert raters, comprising photographers and graphics artists, assessed each image at least twice in multiple sessions spanning several days, \hl{resulting in 20 highly reliable ratings per image}. Annotators were rigorously selected based on several metrics, including self-consistency, to ensure their reliability. The dataset includes rich metadata with user and machine-generated tags from over 5,000 categories and popularity indicators such as favorites, likes, downloads, and views. With its unique characteristics, such as its focus on high-quality images, reliable crowdsourced annotations, and high annotation resolution, our dataset opens up new opportunities for advancing perceptual image quality assessment research and developing practical NR-IQA models that apply to modern photos. Our dataset is available at \small{\href{https://database.mmsp-kn.de/uhd-iqa-benchmark-database.html}{\url{https://database.mmsp-kn.de/uhd-iqa-benchmark-database.html}}}.


\end{abstract}


\section{Introduction}
\label{sec:introduction}

\begin{figure}
    \centering
    \begin{subfigure}[]{0.33\textwidth}
        \includegraphics[width=\textwidth]{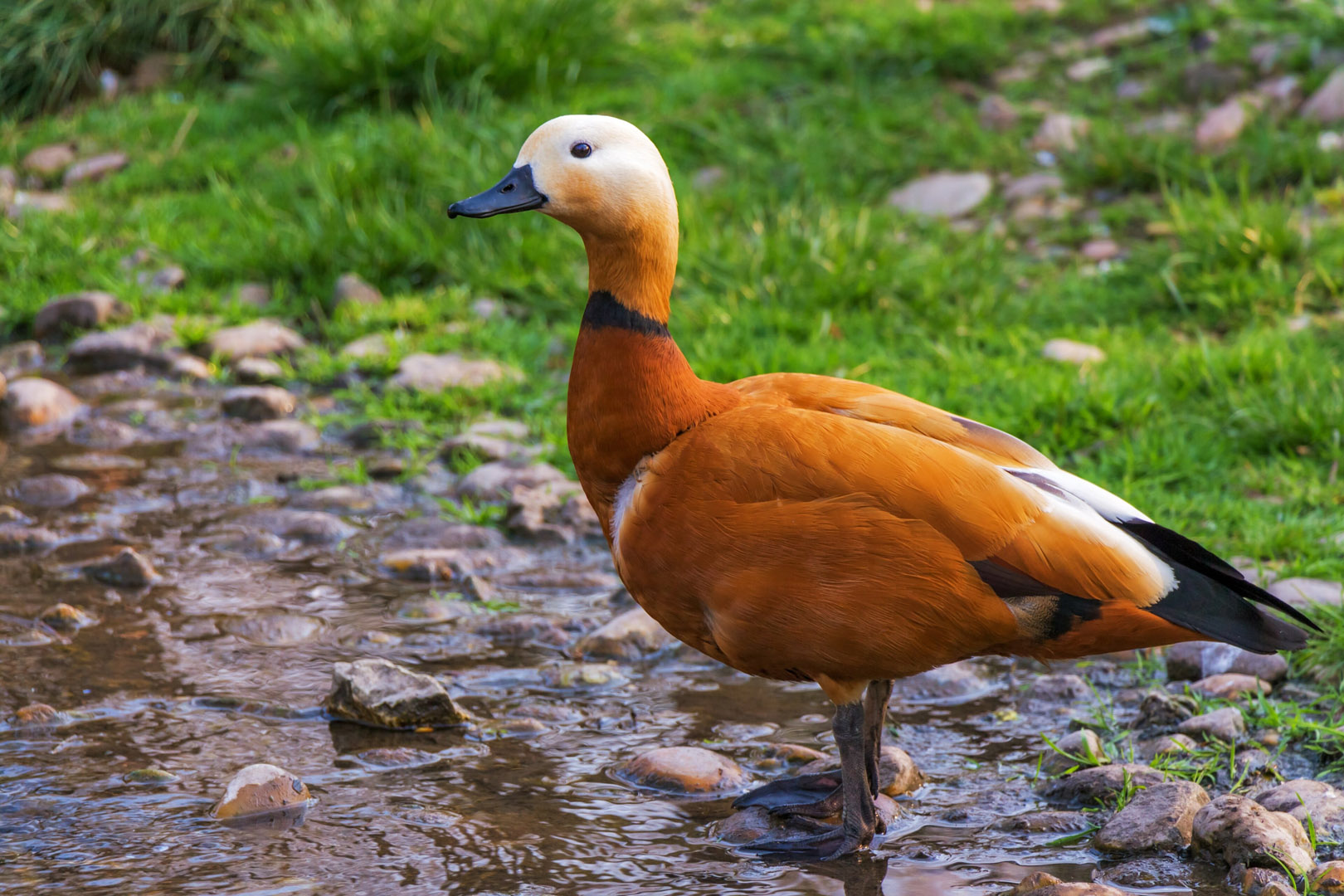}
    \end{subfigure}
    \hspace{-0.5em}
    \begin{subfigure}[]{0.33\textwidth}
        \includegraphics[width=\textwidth]{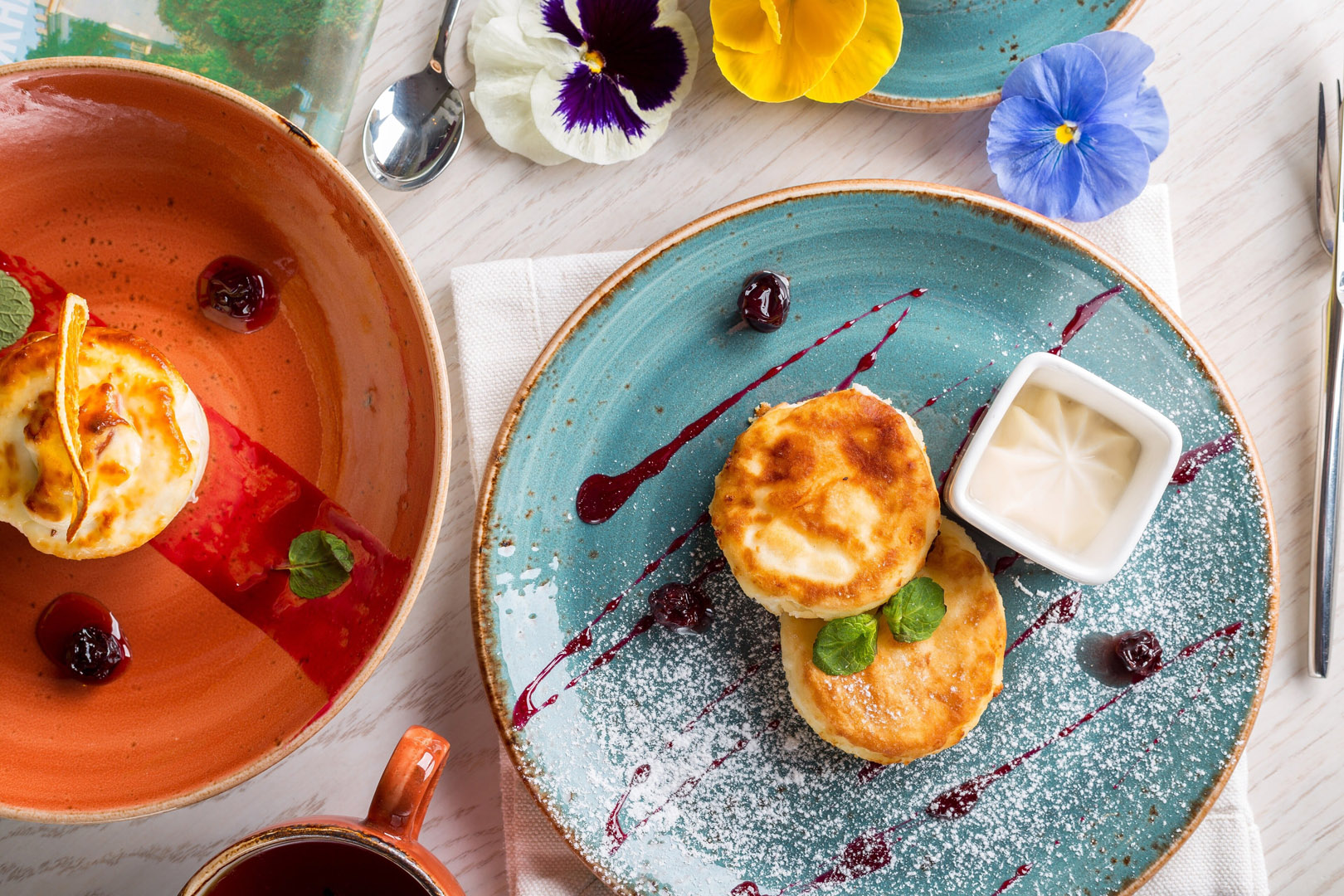}
    \end{subfigure}
    \hspace{-0.5em}
    \begin{subfigure}[]{0.33\textwidth}
        \includegraphics[width=\textwidth]{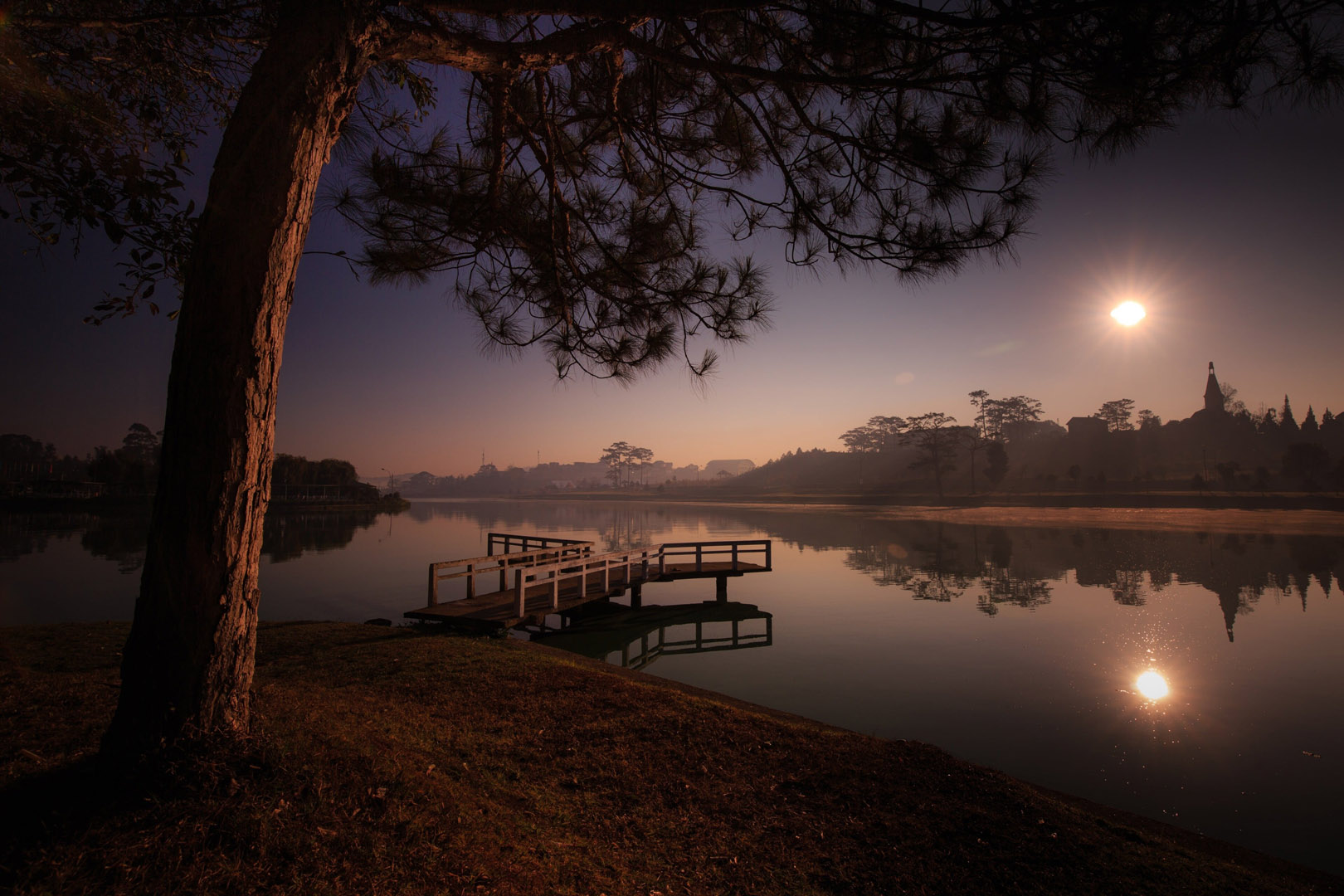}
    \end{subfigure}
    \begin{subfigure}[]{0.33\textwidth}
        \includegraphics[width=\textwidth]{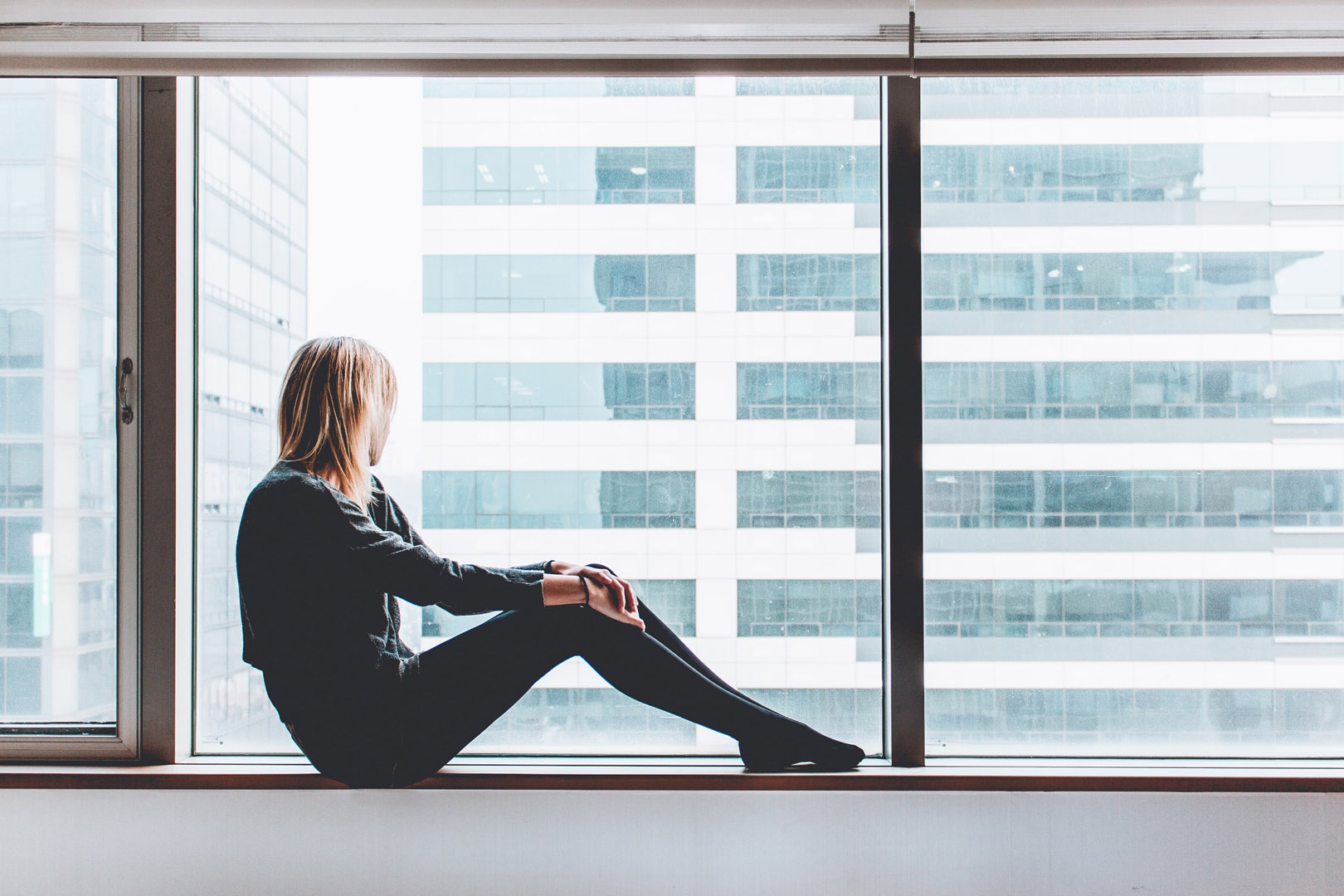}
    \end{subfigure}
    \hspace{-0.5em}
    \begin{subfigure}[]{0.33\textwidth}
        \includegraphics[width=\textwidth]{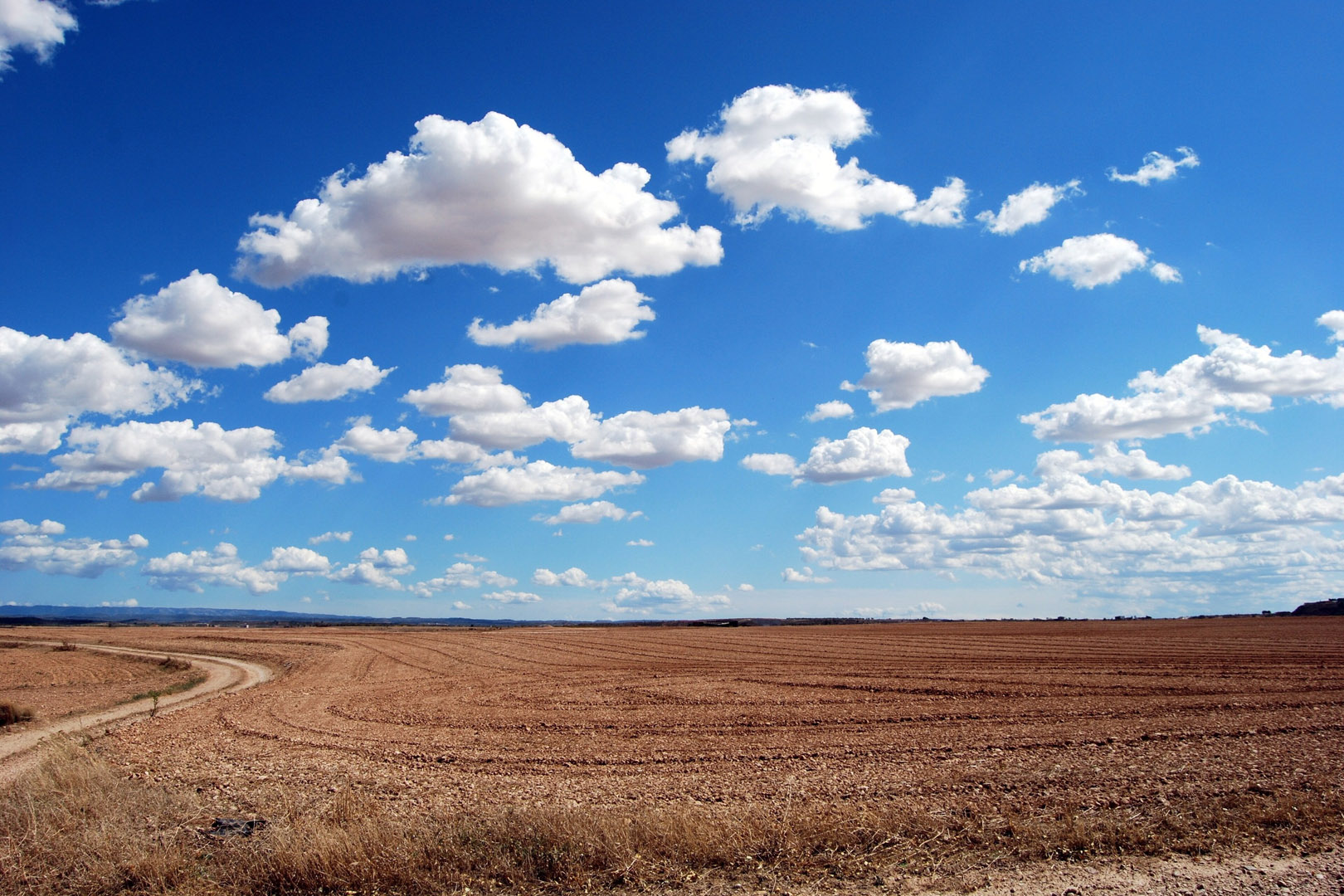}
    \end{subfigure}
    \hspace{-0.5em}
    \begin{subfigure}[]{0.33\textwidth}
        \includegraphics[width=\textwidth]{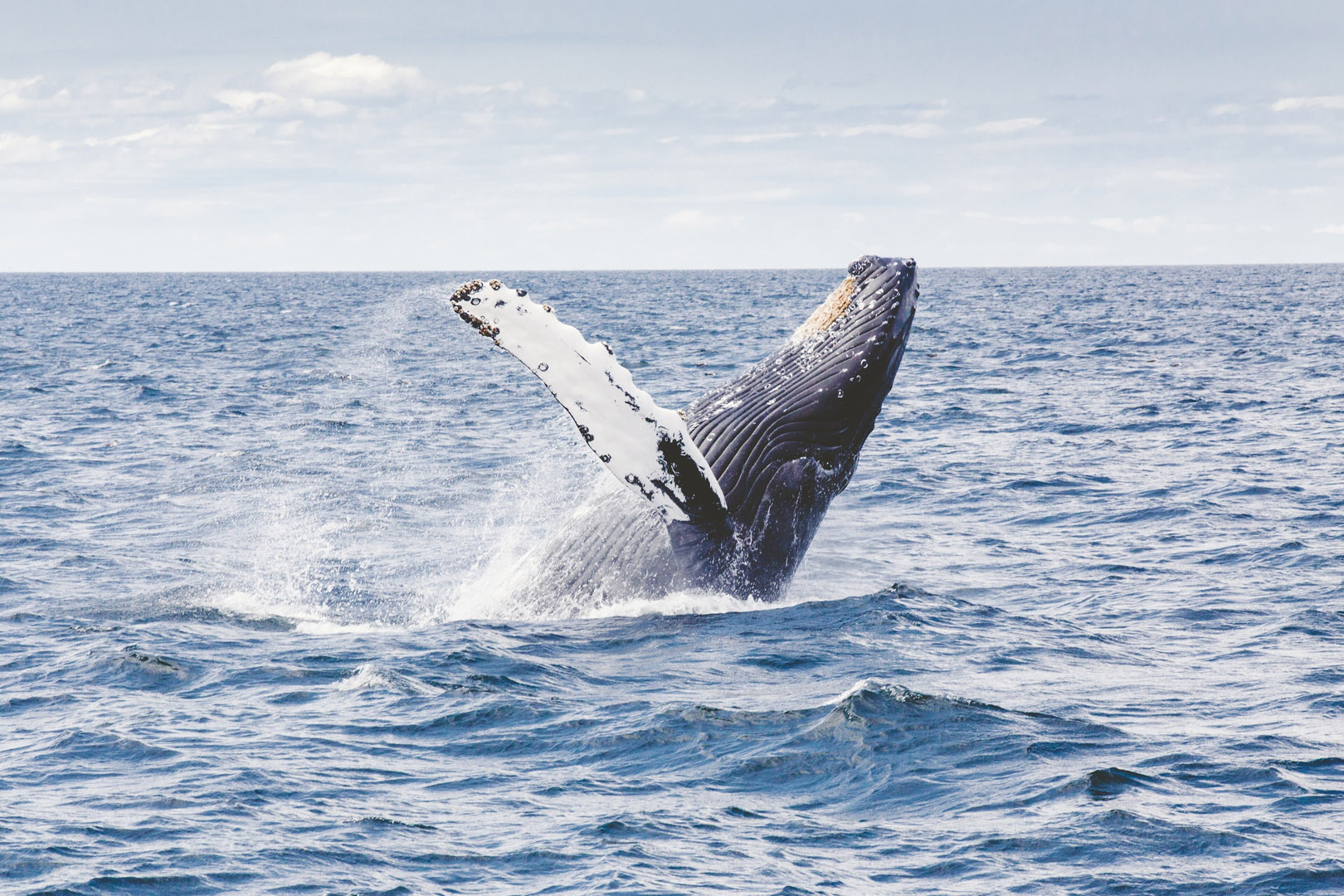}
    \end{subfigure}
    \caption{Sample images from our dataset. The authors of the images are, from left to right, top to bottom: `Bergadder', `Daria-Yakovleva', `Quangpraha', `StockSnap', `pcdazero', and `Free-Photos' from \url{pixabay.com}.}
    \label{fig:teaser_sample_images}
\end{figure}

\subsection{Motivation}
No-Reference Image Quality Assessment (NR-IQA) can revolutionize various applications by automatically evaluating perceptual image quality without requiring a reference. However, NR-IQA models must accurately predict quality and generalize robustly across diverse distortions to realize this potential. Current NR-IQA methods perform well on images with pronounced distortions at Standard Definition (SD) resolutions but struggle on higher-resolution images with subtle quality degradations \cite{wiedemann2023KonxCrossresolutionImageb, huang2024HighResolutionImagec}. Modern cameras commonly capture UHD+ images, on which existing NR-IQA models are inaccurate and inefficient. Moreover, no IQA datasets specifically target high-resolution images in the high-quality range (HR-HQ), which is crucial for discerning subtle degradation. Such fine-grained quality assessment is essential for camera benchmarking, professional-grade photo curation, and optimizing camera parameters. Therefore, to unlock the full potential of NR-IQA, we need novel datasets and models that reliably subjectively assess high-resolution images across the high-quality spectrum.

\subsection{Problem and Challenges}
Only a small fraction of existing IQA datasets are annotated at high resolutions, which is insufficient for training models that generalize well to these resolutions. Furthermore, current datasets are skewed towards average and low-quality images. This under-representation of high-quality images is a typical example of a class imbalance problem, which impairs model performance.

High-resolution and quality IQA datasets are needed to address these issues. However, creating HR-HQ datasets presents two key challenges: annotation costs and reliability. Laboratory studies offer reliable annotations by using high-resolution screens and controlling viewing conditions, but they are limited in scale due to the substantial time commitment required from participants. Crowdsourcing studies are more affordable and scalable but suffer from reduced reliability caused by participants' uncontrolled viewing environments and lower-resolution displays. Consequently, creating large-scale and reliable IQA datasets for high-resolution images is an ongoing challenge.

\subsection{Proposed Approach}
We propose an approach to creating a reliable NR-IQA database for training machine learning models. Our method addresses the limitations of existing datasets by focusing on three key aspects. First, we enhance the reliability of crowdsourcing through an improved user interface that controls display conditions across different screens. Second, we introduce reliability controls that don't require ground truth by leveraging the participants' long-term self-consistency. Third, we engage expert participants to ensure annotation quality.

We source CC0-licensed stock photos from \url{Pixabay.com}, to curate a dataset of high-quality images. However, because a large portion ($\sim$40\%) of the top-ranked images were not authentic but computer-generated or heavily edited, we filtered them out, ensuring our dataset consists of genuine, high-quality photographs. \cref{fig:teaser_sample_images} shows sample images from our dataset.
This approach seeks to overcome the limitations of existing datasets and provide a valuable resource for advancing NR-IQA research and development.

\section{Related Works}
\label{sec:related-works}

We will review the current IQA datasets and quality prediction models, focusing on the research opportunities created by our new dataset.

\subsection{IQA Datasets}
The traditional approach to IQA dataset creation was to collect a set of pristine images and subject them to (mixtures of) \textit{artificial distortions} at multiple distortion levels, as presented in \texttt{LIVE} \cite{sheikh2005live}, \texttt{TID2013} \cite{ponomarenko2015image}, \texttt{CID} \cite{larson2010most} and \texttt{KADID-10k} \cite{lin2019kadid}.

Training machine learning models on specific degradation types can lead to overfitting and poor generalization, particularly regarding \textit{authentically distorted} images. These images feature a complex mixture of real-world distortions common on online photography platforms and social media.

As a result, research has increasingly focused on quality annotated image collections from practical applications. This led to the growth of \textit{authentically distorted} datasets. \texttt{LIVE in the Wild} \cite{ghadiyaram2015massive} highlighted the \textit{authentic} distortion paradigm and demonstrated \textit{crowdsourcing} as a viable alternative to traditional subjective annotation in controlled laboratory settings.

With deep neural networks in mind, \texttt{KonIQ-10k} \cite{hosu2020koniq} was the first larger dataset containing a wide diversity of content for model training. The sampling approach for this dataset aimed to achieve a uniform distribution of selected images across multiple quality-related indicators.

\begin{figure}[t]
    \centering
    \begin{tikzpicture}[scale=0.15]
    \draw [line width=0.2mm] (0, 0) rectangle (16.320, 12.240);
    \draw [line width=0.1mm] (0.5, 0.5) rectangle (4.340, 2.660);
    \draw [fill=black] (1, 1) rectangle (2.024, 1.768);
    \end{tikzpicture}
    \caption{Resolution discrepancy of a $1024\times 768$px/0.7MP image (filled rectangle) as used in many IQA datasets vs. a $3840\times 2160$px UHD/8.3MP image (inner frame) that is common in our new dataset. The outer frame illustrates a $16320\times 12240$px/200MP image captured by recent smartphone sensors, pointing to future challenges.}
    \label{fig:resolution_discrepancy}
 \end{figure}
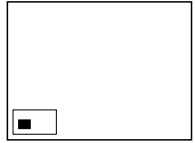

Although existing IQA datasets resemble contemporary images in many respects, their annotated data does not align well with the advancements in modern camera technology, such as improvements in technical quality and resolution, as illustrated in \cref{fig:resolution_discrepancy}.
Image resolution has a measurable effect on perceived quality, as shown in the study presented alongside the \texttt{KonX} \cite{wiedemann2023KonxCrossresolutionImageb} cross-resolution IQA dataset and related \cite{hammou2024effect}. In short, down-scaling has the largest effect on medium-quality images, improving their appearance. This is less the case for the high and low end of the quality scale; thus, excellent and poor quality images are generally unaffected by scaling and stay at their respective level of subjective quality.
Furthermore, the authors \cite{wiedemann2023KonxCrossresolutionImageb} highlight the importance of the reliability of subjective ratings. They argue that consistency in repeated ratings relates to reduced noise in individual annotations. The noise can arise from participants' variable degree of attention, saliency effects, or biases and misinterpreting the scale. Participants who give similar ratings on a second presentation of an image are likely to have lower overall noise. Thus, self-consistency checks are effective. Nevertheless, the annotation and presentation methods in IQA are still evolving with boosted presentation \cite{men2021subjective} and just noticeable differences \cite{lin2022large} as more recent approaches investigating fine differences.

The curse of dimensionality presents a significant challenge in assessing high-resolution images. The increased sparsity of higher dimensions might need more annotated data, straining computational resources and limiting the generalization of IQA methods. Researchers have developed datasets that link global image quality with local patch-based assessments to address this. The concept of patch-wise quality was introduced in \texttt{KonPatch-30k} \cite{wiedemann2018disregarding} and expanded in \texttt{Paq-2-Piq} \cite{ying2020patches}. In addition, distortion-strength labeled datasets like \texttt{KADIS-700k} \cite{lin2019kadid} enhance the generalization of IQA models via self-supervision \cite{madhusudana2022image,agnolucci2024arniqa}. These datasets might lead to new methods that work well on high-resolution images.

Among existing datasets, the most similar to ours is the \texttt{HRIQ} \cite{huang2024HighResolutionImagec} dataset. It comprises 1120 images with a size of $2880\times2160$px annotated in a lab study, with participants primarily recruited from university students. Our proposed dataset improves over \texttt{HRIQ} in the number of images and their resolution. Moreover, our annotators were selected from photographers/graphics professionals who work with high-resolution screens in their conventional environments. This improved rating reliability.

A comparison of existing IQA datasets is presented in \cref{tab:iqa_datasets}. The reliability of annotations varies according to the strategies used, influenced by the participants' motivation and commitment, which is often tied to the level of compensation.

At one end, we have high-commitment and high-compensation cases, such as the freelancer participants in KonX \cite{wiedemann2023KonxCrossresolutionImageb}, who were hired for the entire annotation project and rigorously tested. In the middle, datasets like TID \cite{ponomarenko2009tid2008, ponomarenko2015image} were mostly annotated in university settings, supplemented by remote contributions, presumably from students or collaborators. At the other end, lower-paid crowdsourcing participants \cite{lin2019kadid, hosu2020koniq, ying2020patches} handle smaller tasks with less commitment to the overall experiment.

\begin{table}[t]
 	\caption{Comparison of existing IQA datasets based on several characteristics such as the year the dataset was released, number of source/reference images, distortion types and levels (for each type), total number of distorted images, annotation strategy, and image resolutions.}
        \centering
        \resizebox{\linewidth}{!}{ 
	\begin{tabular}{lccccccc}
		\toprule
		Database 								& Release~~  & \makecell{Source\\Images} & \makecell{Distortion\\Types} & \makecell{Distortion\\Levels} & \makecell{Distorted\\Images} 	& \makecell{Annotation\\Strategy} 	    & \makecell{Image\\Resolution (px)} 	    \\
		\hline                                          
		LIVE \cite{sheikh2005live}			    & 2006  & 29  						&  5							&								& 799							& lab study 							& mostly $768\times512$ 		     \\
		TID2008 \cite{ponomarenko2009tid2008}	& 2008  & 25						& 17							& 4								& 1700							& remote + lab 			                & $512 \times 384$			     \\
		CSIQ \cite{larson2010most}    			& 2010  & 30 						&  6 							& 4 or 5 						& 866 							& lab study 							& $512  \times 512$  			     \\
		TID2013 \cite{ponomarenko2015image}		& 2013  & 25						& 24							& 5								& 3000							& remote + lab			                & $512 \times 384$			     \\
		CID2013 \cite{virtanen_cid2013_2015} 	& 2013  & 480						& (12-14 cameras)				& --							& --							& lab study							    & $1600 \times 1200$	 \\
        MDID \cite{sun2017mdid} & 2017 & 20 & 5 & 4 & 1600 & lab study & $512\times 384$\\
        KonIQ-10k \cite{hosu2020koniq}			& 2018  & 10073						& --							& --							& --							& crowdsourcing						    & $1024 \times 768$ 			     \\
		KADID-10k \cite{lin2019kadid} 			& 2019  & 81 						& 25 							& 5 							& 10125 						& crowdsourcing 					    & $512 \times 384$ 			     \\
		SPAQ \cite{fang2020perceptual} & 2020 & 11125 & (66 smartphones) & -- & -- & lab study & shorter side 512px\\ 
		PaQ-2-PiQ \cite{ying2020patches}		& 2020  & 39810						& -- 							& --							& --							& crowdsourcing						    & variable						     \\
        KonX \cite{wiedemann2023KonxCrossresolutionImageb} &2023 & 420 & -- & -- & -- 	& freelancers 	& max $2048 \times 1536$ \\
        HRIQ \cite{huang2024HighResolutionImagec}  & 2024 & 1120 & -- & -- & -- 	& lab study 	& $2880\times2160$ 	\\ \midrule
        \textbf{UHD-IQA (Ours)} & 2024 & 6073 & -- & -- & -- & freelancers & mostly $3840 \times 2160$ \\  
        
		\bottomrule
	\end{tabular}}
\label{tab:iqa_datasets}
\end{table}

\subsection{Auxiliary Image Datasets}
Beyond IQA, several other image databases \cite{murray2012ava, li2023LSDIRLargeScale, zhang2021benchmarking, liu2020comprehensive} might help create predictive models. For instance, one way to do this is to pre-train an image encoder in a self-supervised manner \cite{madhusudana2022image, agnolucci2024arniqa, agnolucci2024qualityaware}.

Alternatively, one could utilize datasets annotated for related tasks. An example is the \texttt{AVA} \cite{murray2012ava} dataset, which assesses image aesthetics rather than their technical quality. Nonetheless, the two tasks are different, and training on aesthetics-related labels could only be marginally beneficial for predicting technical quality. 

Image restoration datasets \cite{li2023LSDIRLargeScale, zhang2021benchmarking, liu2020comprehensive} might also be suited for pre-training. For instance, \texttt{LSDIR} is a large-scale dataset for image restoration comprising about 87,000 high-resolution images collected from Flickr. Human annotators manually inspected these images to ensure high quality. Zhang \etal \cite{zhang2021benchmarking} introduce the \texttt{UHDSR4K} and the \texttt{UHDSR8K}, which contain 8099 images of 4K resolution and 2966 images of 8K resolution. The images in these datasets are collected from the Internet and depict a broad range of content.



\subsection{NR-IQA Models}
In recent years, no-reference image quality assessment has drawn significant interest \cite{su2020blindly, li2020norm, wiedemann2023KonxCrossresolutionImageb, madhusudana2022image, agnolucci2024arniqa, wang2023exploring, agnolucci2024qualityaware}, given its practical applications in both research and industry settings.
Supervised learning has proven to be an effective technique for NR-IQA, as demonstrated by the performance of several methods relying on it \cite{su2020blindly, li2020norm, wiedemann2023KonxCrossresolutionImageb}. For instance, HyperIQA \cite{su2020blindly} introduces a self-adaptive hypernetwork that separates content understanding from quality prediction. The model is trained on $224\times224$px image patches by minimizing the $L_{1}$ loss between the predicted and ground truth quality scores.
The Effnet-2C-MLSP \cite{wiedemann2023KonxCrossresolutionImageb} model combines the output of two multi-level spatially pooled \cite{hosu2019effective} EfficientNet-B7 \cite{tan2019efficientnet} feature extractors operating on different input image resolutions, which achieves excellent performance in cross-resolution quality assessment.

A different line of research is based on pre-training a self-supervised image encoder and then learning a linear regression using the labeled MOS \cite{madhusudana2022image, agnolucci2024arniqa}. For example, ARNIQA \cite{agnolucci2024arniqa} pre-trains the encoder by maximizing the similarity between different images degraded by a similar procedure. Hence, the encoder learns to generate similar representations for images with similar distortion patterns, regardless of their content.

Recently, several works \cite{wang2023exploring, zhang2023blind, agnolucci2024qualityaware} have tackled NR-IQA by relying on vision-language models, such as CLIP \cite{radford2021learning}. CLIP-IQA \cite{wang2023exploring} employs an out-of-the-box CLIP model to compute the quality score by measuring the similarity between the image and two antonym prompts, such as ``Good/Bad photo". CLIP-IQA$+$ \cite{wang2023exploring} additionally learns the antonym prompts using the labeled MOS. In contrast, QualiCLIP \cite{agnolucci2024qualityaware} improves over CLIP-IQA by employing a self-supervised quality-aware image-text alignment strategy to make CLIP generate representations that correlate with the intrinsic quality of the images.

Despite the many approaches and techniques presented above, none of these methods are designed for high-resolution images. As a result, it is challenging to fully utilize the detailed information available in UHD images. We hope that the release of our dataset will foster research on approaches specifically tailored for handling high-resolution images effectively and efficiently.


\section{Database Sampling}

The images composing our dataset were sampled from Pixabay\footnote{\hyperref[https://pixabay.com]{\url{https://pixabay.com}}}. We indexed an initial collection of images classified as photos on the platform, amounting to about 150,000 images of resolutions greater than UHD-1 ($3840 \times 2160$px). These were sorted by normalized favorites \cite{wiedemann2023KonxCrossresolutionImageb}, and the top 10,000 were selected for further sampling. \hl{From these, synthetic images were removed in a subjective study. Three participants were tasked with individually assessing each image to identify those that appeared synthetic, such as computer-generated renderings or drawings. They were asked to focus particularly on images that closely resemble real photographs but are likely not authentic. Once at least one participant identified these images, they were filtered out.}

\begin{figure}[h]
    \centering
    \begin{subfigure}[b]{0.3\textwidth}
        \centering
        \includegraphics[width=\textwidth]{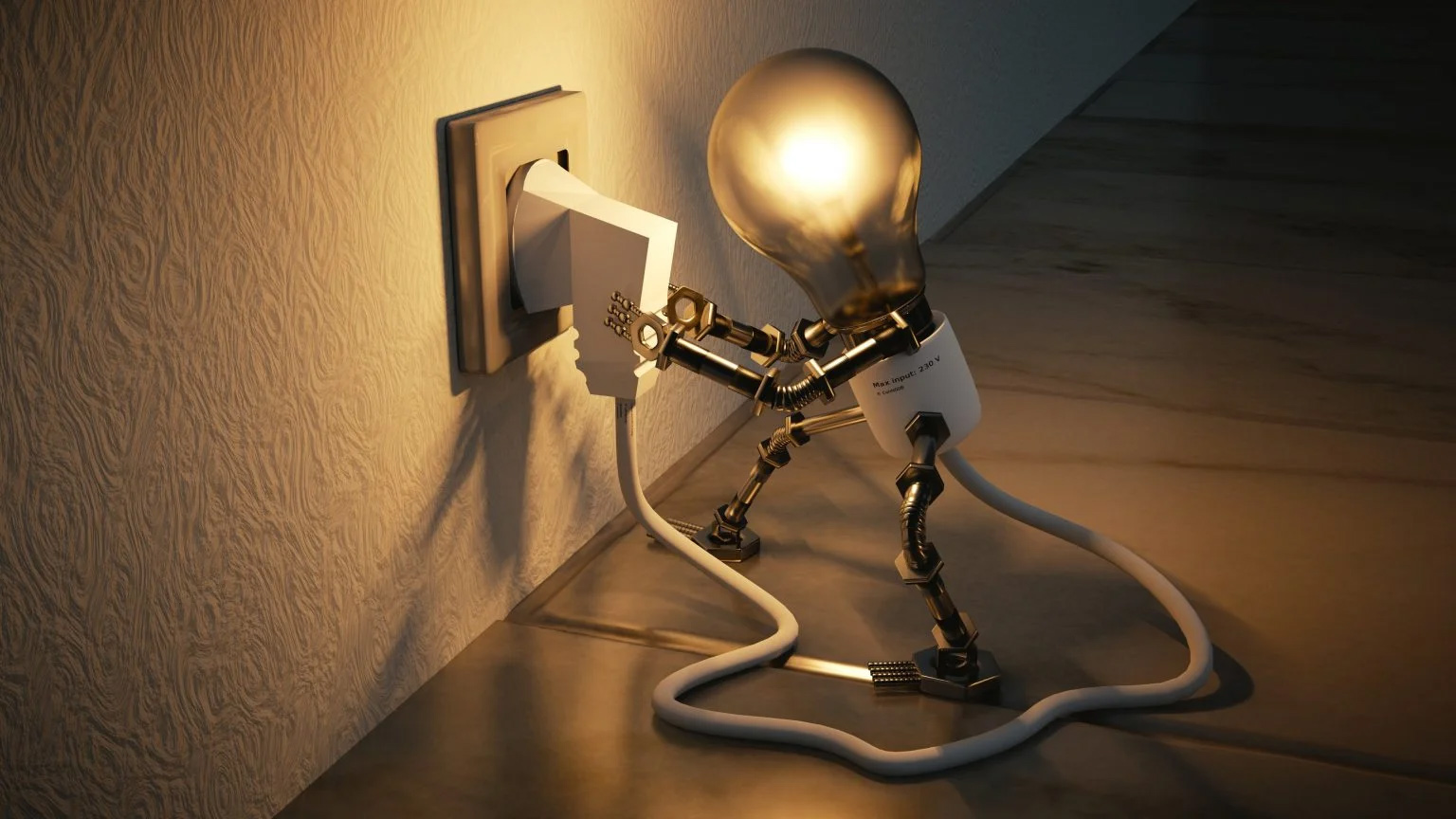}
        \caption{Graphic renderings}
    \end{subfigure}
    \hfill
    \begin{subfigure}[b]{0.3\textwidth}
        \centering
        \includegraphics[width=\textwidth]{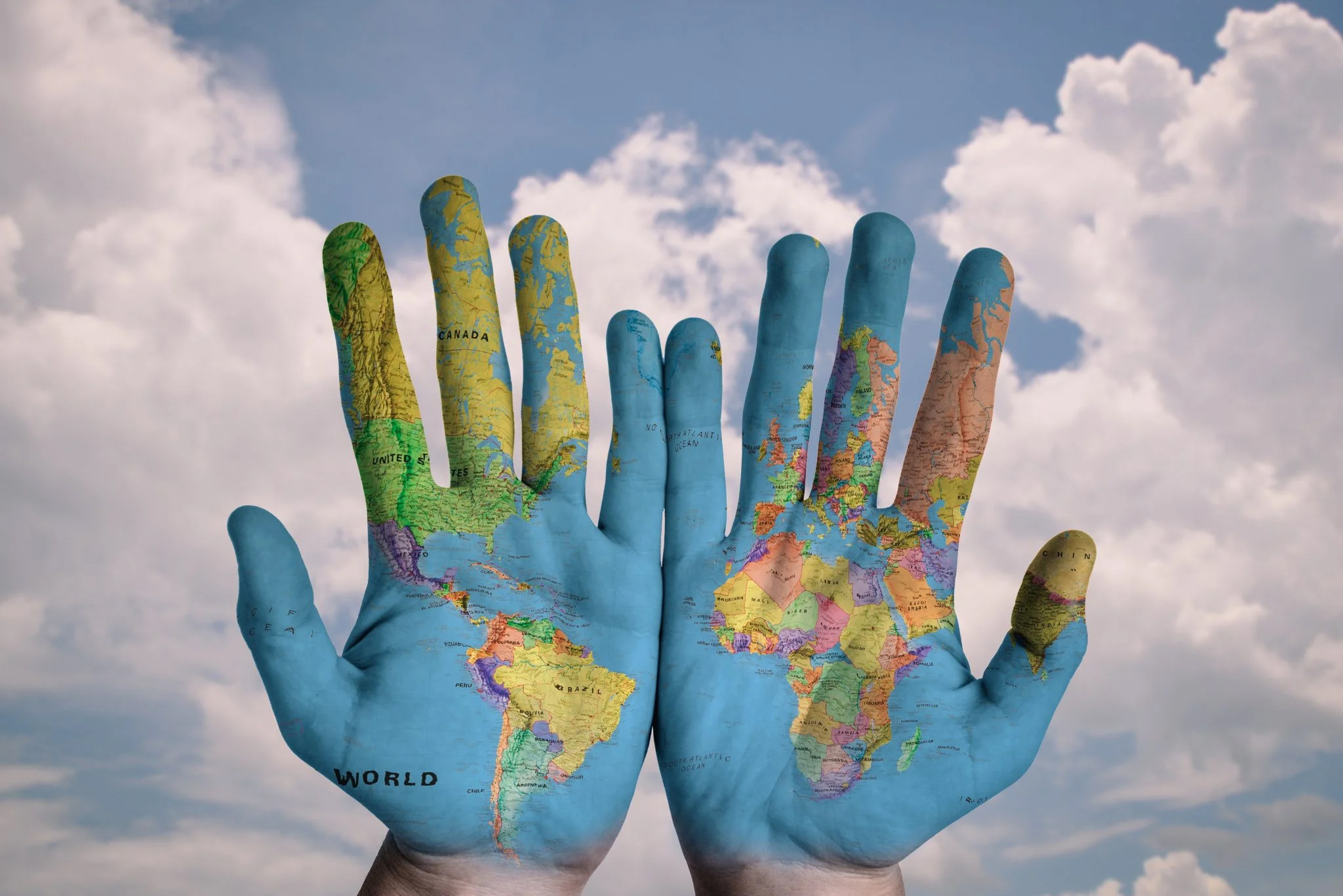}
        \caption{Photo composites}
    \end{subfigure}
    \hfill
    \begin{subfigure}[b]{0.3\textwidth}
        \centering
        \includegraphics[width=\textwidth]{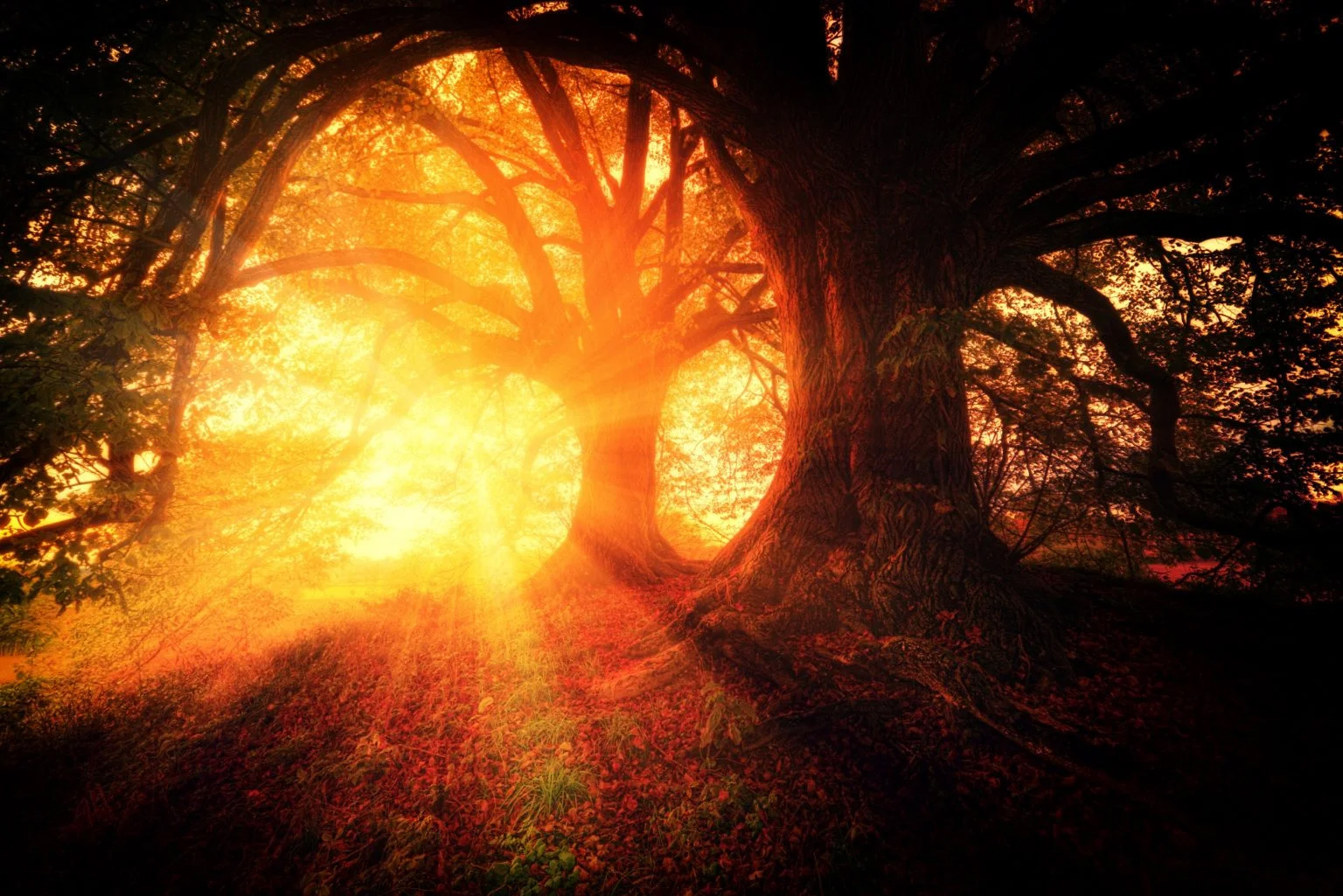}
        \caption{Overly edited}
    \end{subfigure}
    \caption{Examples of synthetic images removed from the initial collection, leaving only authentic photos in the annotated dataset. The authors of the images are, from left to right: `ColiN00B', `stokpic', and `jplenio' from \url{Pixabay.com}.}
    \label{fig:synthetic}
\end{figure}

For photo composites, participants had to exclude images that contained elements that could not exist in real life. Additionally, they had to identify images that have been overly or poorly edited, where the edits resulted in unrealistic or implausible visuals, such as extreme color adjustments or exposure enhancements. We show examples of images removed from the initial collection in \cref{fig:synthetic}. The remaining 6,073 images constitute our proposed dataset.

\section{Subjective Study}
\label{sec:subjective_study}
To collect an IQA database, we conducted a subjective study using the IQAvi web application \cite{wiedemann2023KonxCrossresolutionImageb}. Participants were presented with a series of images and asked to rate their quality on a scale from "bad" (1\%) to "excellent" (100\%). \hl{Each of the ten participants rated every image twice, resulting in 20 ratings per image.}
\hl{Most of the original images from} \url{pixabay.com} \hl{had resolutions higher than UHD-1. To ensure the images were not too large for participants' screens, before the annotation process, we downscaled them to a fixed width of 3840 pixels using Lanczos interpolation.} The height was proportionally scaled to maintain the original aspect ratio. Thus, the height of some images is smaller than the standard 2160 pixels. \hl{This approach allowed us to meet several criteria: maintaining manageable image sizes, ensuring consistency across the dataset, and increasing the use of screen space.}

\subsection{Experimental Setup}
\label{subsec:experimental_setup}
Participants were required to use the Chrome browser on a display with a diagonal larger than 14 inches and a native resolution higher than $2560\times 1440$ pixels. The browser was set to full-screen mode during the experiment, and participants were instructed to adjust their screen brightness and contrast to see image details clearly. A fast internet connection (5+ Mbps) was recommended to minimize image-loading delays.
Before the main experiment, participants completed a training session to familiarize themselves with the quality rating scale and the specific image defects they should evaluate. \hl{Quality ratings were assigned using a 100-point slider.} The training session included images with gold-standard quality ratings to help participants calibrate and anchor their judgments. The ground-truth images and ratings were sourced from $2048\times1536$ pixel images from the KonX database \cite{wiedemann2023KonxCrossresolutionImageb}.

\hl{Participants employed the panning feature of the annotation tool to view the entire image when it exceeded the screen size. They were explicitly instructed to distinguish between resolution and quality, ensuring their ratings were based on the degree of degradation rather than the display resolution, regardless of the image size that fit their screens. Previous studies, such as} \cite{goring2023QualityAssessmentHighera}\hl{, have shown that cropping does not significantly impact quality assessment. However, to further minimize potential biases, we verified through experiment logs that participants actively panned and explored images before assigning ratings.}


\subsection{Quality Assessment Criteria}
\label{subsec:quality_assessment_criteria}
Participants were instructed to assess the technical quality of images based on the level of \textit{annoyance caused by visible defects}, such as noise, blur, compression artifacts, and color distortions. They were informed that technical quality is distinct from aesthetic appeal or attractiveness and that images with high technical quality are not always highly aesthetically pleasant.
However, participants were also advised that in certain cases, such as macro or close-up photography, some defects, such as background blur (``bokeh''), may be an intrinsic part of the composition and should only be considered as quality-degrading when they become annoying to the observer.

\subsection{Participant Selection}
\label{subsec:participant_selection}
\hl{To ensure the reliability of the subjective study, we carefully selected participants, inviting freelancers with verified expertise in the visual arts. Many of these participants had substantial experience in quality assessment as part of their professional activities, further supported by their impressive track record on Freelancer.com. On average, they have received 52 ratings in previous projects, all maintaining a perfect 5/5 average rating, except for one participant who had 4.9/5. The participants' expertise spans several professional categories: 4 are photographers, 3 specialize in video production and editing, 5 are graphic designers, and one has experience in printing. Some participants possess skills in multiple categories. 

Prior studies have highlighted notable differences between experts (e.g., graphic design professionals) and novices (general crowd workers) in the context of NR-IQA} \cite{hosu2018expertise}\hl{, with findings indicating that experts tend to provide less biased and more reliable ratings. Additionally, the freelancer's commitment to the task was supported by their professional reputation and our direct communication channel throughout the study. Post-completion performance reviews impacted their long-term freelancer profiles, reinforcing their dedication to the job. Overall, participants were both capable and motivated, ensuring high-quality results.}

\hl{The participant selection process included an initial training phase, followed by using hidden gold-standard questions. The questions presented participants with images of known ground-truth quality scores, which were randomly interspersed among the regular images to be annotated.} The ground-truth quality scores were based on the Mean Opinion Scores (MOS) of images from the \texttt{KonX} database, with an acceptable range of $\pm$ 1 standard deviation. Participants' responses were considered correct if they fell within these predefined intervals.
\hl{Of the 25 initial contestants, 13 met the accuracy and correlation requirements on the gold-standard test images. The requirements were having an accuracy greater than 70\% and a Spearman's Rank-order Correlation Coefficient (SRCC) greater than 0.75 relative to the ground-truth MOS.} A visual inspection of the distribution of scale ratings was made to check that the full range of the 100-point scale was used as uniformly as the dataset allowed, e.g., it did not have peaks at the absolute category ratings (ACR) tick-marks.

\hl{However, two participants were later removed from the final dataset due to incomplete submissions and another due to unreliable performance during the main study. This resulted in 10 reliable participants whose responses were used for the analysis. The minimum achieved accuracy on the hidden test images was 75\%, and the minimum SRCC was 0.85. This was calculated over 970 test images.}

\hl{The stringent participant selection process, which included training, testing, and ongoing performance monitoring, aimed to ensure that the collected subjective quality scores were reliable and accurate. Moreover, the user interface and controls were meticulously designed to maximize the experiment's reliability. Given these factors, this remote study provides a robust framework for remotely gathering accurate subjective quality ratings.}

\subsection{Annotation Procedure}


The IQA dataset comprises 6073 images, which were divided into 61 batches or subsets. Each batch was presented twice to each participant, organized into two annotation rounds: the first presentation (batches 1 to 61) and the second (batches 62 to 122). Each annotation round took several days for each participant to complete. During each presentation, the order of images within each batch was randomized. Additionally, the sequence of batches was randomized uniquely for each participant, ensuring a fixed yet individualized list of batches to annotate.

To mitigate the risk of participants memorizing their responses, the second presentation of any given batch was temporally separated from the first. This separation was achieved by first assigning a random order to the first presentation of the 61 batches. For the second round, we took the first half of the batches from the first round, randomized them again, and concatenated them with the randomized second half of the batches from the first round. This method guaranteed that there were always at least 30 batches between the two presentations of the same subset of images, ensuring sufficient separation of at least a few days to prevent memorization.

\hl{In the end, we obtained 20 ratings per image (10 annotators $\times$ 2 judgments each), thus meeting the minimum requirement of 15 opinions per stimulus recommended by} \cite{RecommendationITUT913}\hl{. This ensures the reliability of the MOS of the proposed dataset.}

\section{Discussion}

\subsection{Analysis}

\begin{figure}
    \centering
    \begin{subfigure}{0.475\linewidth}
    \centering
        \begin{tabular}{lc}
        \toprule
        Subset & $\#$ of images \\ \midrule
        Training & 4269 \\		
        Validation & 904 \\
        Test & 900 \\ \midrule
        Total & 6073 \\ \bottomrule
    \end{tabular}

        \vspace{27pt}
        \caption{}
        \label{tab:dataset_statistics} 
    \end{subfigure}
    \hfill
    \begin{subfigure}{0.475\linewidth}
        \includegraphics[width=\linewidth]{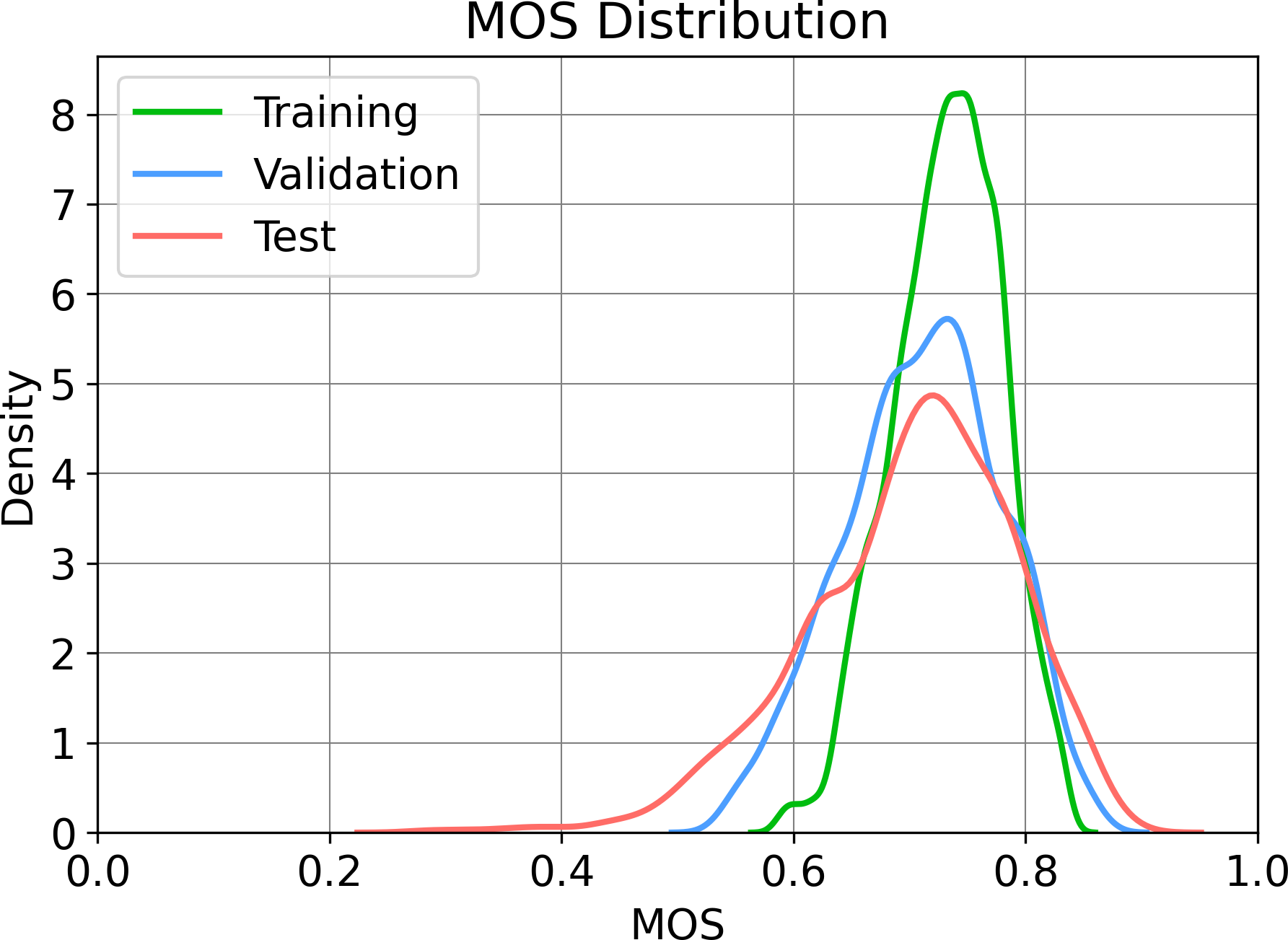}
        \caption{}
        \label{fig:mos_distribution}
    \end{subfigure}
    \caption{(a): Number of images in the training, validation and test set. (b): Distribution of the MOS in each subset.}
    \label{fig:gmad_grepq}
\end{figure}

We aimed to create a benchmark dataset split that will be used to measure the performance of machine learning models in a representative and challenging manner. Models achieving high performance on our test set should suit contemporary, high-quality, high-resolution images. This requires the test subset to be the most diverse.
We divided the dataset into three subsets: approximately 70\% for training, 15\% for validation, and 15\% for testing. The exact number of images in each subset is shown in \cref{tab:dataset_statistics}. First, we selected the test set using a stratified sampling strategy to ensure that the Mean Opinion Scores (MOS) distribution more closely resembles a uniform one \cite{wiedemann2023KonxCrossresolutionImageb}. Next, we selected the validation set from the remaining images and followed a similar sampling procedure. Finally, the remaining examples constituted the training set. Consequently, the test and validation sets exhibit a broader range of MOS values than the training set, as illustrated in \cref{fig:mos_distribution}. Note that when referencing a subset or dataset split, we refer to the official splits introduced in this benchmark.

\subsubsection{Measures of Annotation Reliability}

While the images were rated on a scale of $[1,100]$, the following calculations use a normalized Mean Opinion Score (MOS) that is based on rescaled ratings to $[0,1]$.
We compared the MOS from the first and second rounds to assess the experiment's reliability or repeatability. On the test set, this yielded an SRCC of 0.93 and a Root Mean Square Difference (RMSD) of 0.03, indicating a high consistency between the two annotation rounds.

\begin{figure}
  \centering
  \begin{subfigure}{0.49\linewidth}
    \centering
    \includegraphics[width=\linewidth]{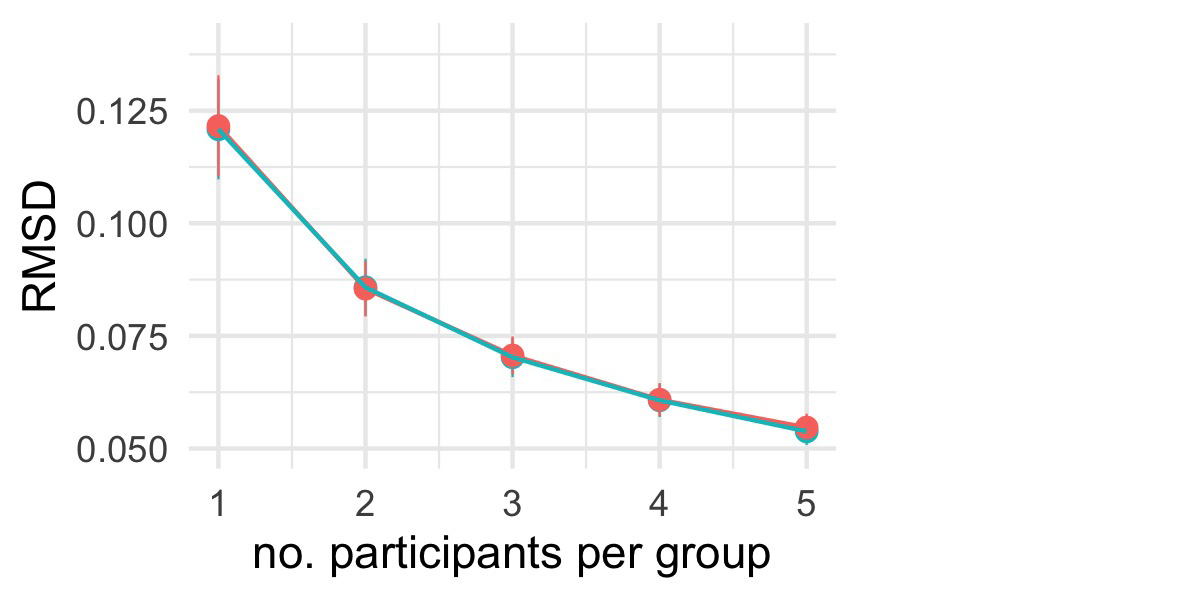}
    \caption{}
    \label{fig:inter-group-agreement-RMSE}
  \end{subfigure}
  \hfill
  \begin{subfigure}{0.49\linewidth}
    \centering
    \includegraphics[width=\linewidth]{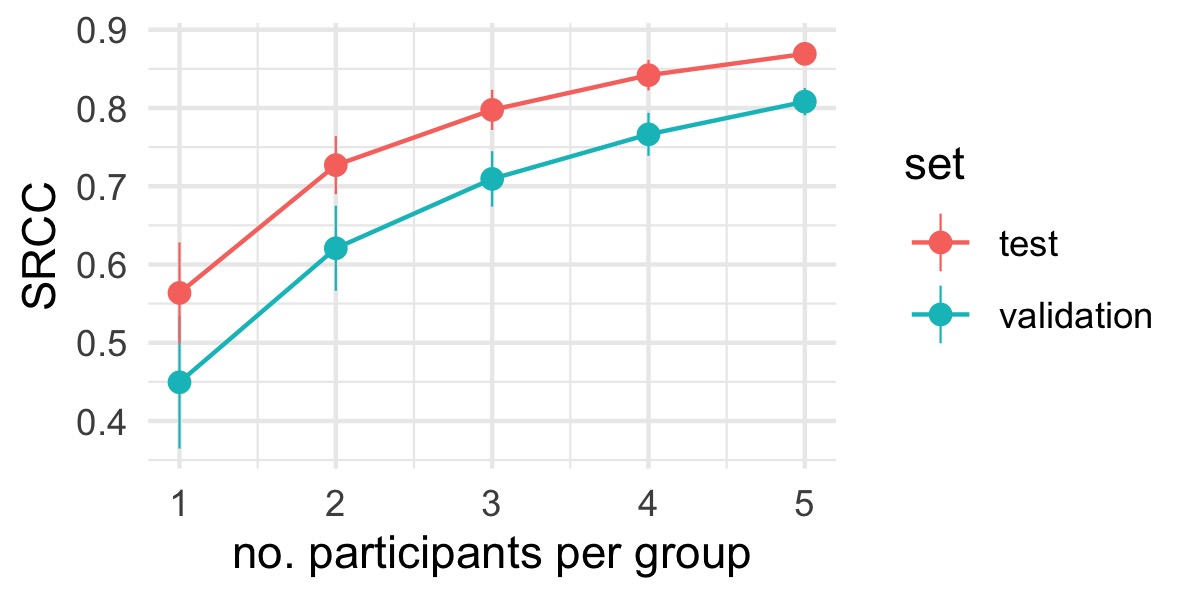}
    \caption{}
    \label{fig:inter-group-agreement-SRCC}
  \end{subfigure}
  \caption{(a): Root Mean Squared Difference (RMSD) between the MOS of different sized groups of participants on the test and validation sets. (b): SRCC between MOS of groups. The error bars show $\pm$1 standard deviation of the performance metrics, and the dots represent the averages. 200 samples of pairs of groups of non-overlapping participants were randomly drawn to compute the statistics.}
  \label{fig:inter-group-agreement}
\end{figure}

Another measure of reliability or repeatability of the experiment was the agreement between groups of participants. We sampled equal-sized non-overlapping groups of participants and compared their MOS. The group's size went up to 5 from a total of 10 participants. This group-wise MOS comparison provides a lower bound for the overall reliability of 10 vs. 10 participants, which is our primary interest. The results for the test set, illustrated in \cref{fig:inter-group-agreement}, indicate an expected SRCC of at least 0.87 and an RMSD of at most 0.055 for the 10 vs. 10 participant comparison. The SRCC for the validation set is slightly lower at 0.81 due to the stratified sampling strategy of the subsets; however, the validation RMSD is 0.054, almost the same as for the test set.

\subsubsection{Implications for Model Performance}

The comparisons between virtual groups of 10 vs. 10 participant agreement and the first vs. second round MOS offer insights into the expected performance of predictive models. While models have the potential to ``denoise`` the original ratings and thereby achieve higher performance relative to the noisy ground truth, their success depends on their generalization capabilities. However, as we will see in the next section, baseline models have not yet reached the precision observed in the MOS derived from groups of 5 participants.


\subsection{Model performance evaluation }
We investigate the performance of several state-of-the-art NR-IQA methods on our dataset. To provide a comprehensive analysis, we consider models based on a broad range of approaches, such as standard supervised learning \cite{su2020blindly, wiedemann2023KonxCrossresolutionImageb}, self-supervised learning \cite{madhusudana2022image, agnolucci2024arniqa}, and vision $\&$ language \cite{wang2023exploring, agnolucci2024qualityaware}. For a fair comparison, we train each model on the training split of our dataset. In particular, for methods based on self-supervised learning and vision $\&$ language, we keep the encoder weights frozen and learn a linear regressor and the prompts \cite{wang2023exploring}, respectively.
We train Effnet-2C-MLSP using images at $512\times384$px and $1024\times768$px resolutions. These are generated by down-scaling $3840\times2160$ crops from our dataset, which are symmetrically zero-padded for those images with insufficient height, which replicates the training approach from the original paper. For the other models, we use (zero-padded) square patches of size $2160$px, except for HyperIQA \cite{su2020blindly}, which employs $224$px crops.

We employ multiple metrics for performance evaluation, namely Kendall's Rank Correlation Coefficient (KRCC), Pearson’s Linear Correlation Coefficient (PLCC), Spearman’s Rank-order Correlation Coefficient (SRCC), Root Mean Squared Error (RMSE), and Mean Absolute Error (MAE). We do not apply logistic regression to the predictions before computing the PLCC. 
Additionally, we assess the computational efficiency of the models by measuring the number of Multiply-Accumulations operations (MACs)\footnote{We rely on the \texttt{ptflops} library for measuring the MACs.} required for a forward pass with an input image size of $3840\times2160$ pixel.

We report the results on the validation and test set in \cref{tab:validation_model_performance} and \cref{tab:test_model_performance}, respectively. Based on their performance metrics, the overall method ranking is similar between the validation and test subsets. However, due to the sampling strategy, the test set contains a wider MOS range, leading to higher correlations. 

Regarding individual methods, first, we observe that HyperIQA \cite{su2020blindly} achieves significantly lower performance than the other baselines. We attribute this outcome to the fact that HyperIQA computes the final quality score as the average of 25 different $224\times224$ random patches extracted from the input image. In contrast, the other methods take the whole image as input and output a single quality score. Even though the strategy adopted by HyperIQA leads to considerably fewer MACs, it does not take full advantage of the amount of information contained in high-resolution images. Effnet-2C-MLSP aggregates features from downscaled inputs at different resolutions using a two-stream MLSP extractor backbone. This results in predictions that are more correlated with the ground truth than in the case of HyperIQA. We assume that the high-resolution inputs impose a large domain shift from the ImageNet pretraining. Thus, we expect the model to perform better with additional training data.
Furthermore, we notice that CLIP-based models, namely CLIP-IQA+\cite{wang2023exploring} and QualiCLIP\cite{agnolucci2024qualityaware}, obtain the best results according to the correlation-based metrics (KRCC, PLCC, and SRCC) metrics but fall behind when considering the error-based ones (RMSE and MAE). We suppose this is because CLIP-IQA+ and QualiCLIP compute the quality score based on the cosine similarity between the input image and two antonym prompts. Due to the way CLIP is trained \cite{radford2021learning}, this cosine similarity can span the entire $[0, 1]$ range. Therefore, the predicted quality scores cover a broader range than the one of the ground-truth MOS (see \cref{fig:mos_distribution}), leading to higher absolute errors. Finally, we observe that the self-supervised learning baselines (\ie CONTRIQUE \cite{madhusudana2022image} and ARNIQA \cite{agnolucci2024arniqa}) achieve the most balanced performance, as they obtain good results for all the metrics while requiring fewer MACs than CLIP-based models. 

We are only considering performance in a broad sense. We have not studied the generality of the baseline models outside the scope of the database and its inherent biases. For instance, we have not analyzed model fairness relative to the different characteristics of the images and the subjects depicted. The preferences towards certain image types of the \url{Pixabay.com} community determine inherent biases that can exist in the trained models. This could lead to spurious correlations between the types of subjects and quality levels.



\begin{table}[]
    \caption{Evaluation of the performance of the baselines on the validation set. $\uparrow$ means that higher values are better, $\downarrow$ means that lower values are better. The best and second-best scores are highlighted in bold and underlined, respectively.}
    \centering
    \begin{tabular}{lcccccc}
        \toprule
        Method & KRCC$\:\uparrow$ & PLCC$\:\uparrow$ & SRCC$\:\uparrow$ & RMSE$\:\downarrow$ & MAE$\:\downarrow$ & MACs (G)$\:\downarrow$ \\ \midrule
        HyperIQA \cite{su2020blindly} & 0.359 & 0.182 & 0.524 & 0.087 & 0.055 & \textbf{211} \\		
        Effnet-2C-MLSP \cite{wiedemann2023KonxCrossresolutionImageb} & 0.445 & 0.627 & 0.615 & 0.060 & 0.050 & \underline{345} \\
        CONTRIQUE \cite{madhusudana2022image} & 0.521 & 0.712 & 0.716 & \textbf{0.049} & \textbf{0.038} & 855 \\
        ARNIQA \cite{agnolucci2024arniqa} & 0.523 & 0.717 & 0.718 & \underline{0.050} & \underline{0.039} & 855 \\
        CLIP-IQA+ \cite{wang2023exploring} & \underline{0.546} & \underline{0.732} & \underline{0.743} & 0.108 & 0.087 & 895 \\
        QualiCLIP \cite{agnolucci2024qualityaware} & \textbf{0.557} & \textbf{0.752} & \textbf{0.757} & 0.079 & 0.064 & 901 \\ \bottomrule
    \end{tabular}

    \label{tab:validation_model_performance}
\end{table}
\begin{table}[]
    \caption{Evaluation of the performance of the baseline methods on the test set. $\uparrow$ means that higher values are better, $\downarrow$ means that lower values are better. Best and second-best scores are highlighted in bold and underlined, respectively.}
    \centering
    \begin{tabular}{lcccccc}
        \toprule
        Method & KRCC$\:\uparrow$ & PLCC$\:\uparrow$ & SRCC$\:\uparrow$ & RMSE$\:\downarrow$ & MAE$\:\downarrow$ & MACs (G)$\:\downarrow$ \\ \midrule
        HyperIQA \cite{su2020blindly} & 0.389 & 0.103 & 0.553 & 0.118 & 0.070 & \textbf{211} \\		
        Effnet-2C-MLSP \cite{wiedemann2023KonxCrossresolutionImageb} & 0.491 & 0.641 & 0.675 & \underline{0.074} & \underline{0.059} & \underline{345} \\
        CONTRIQUE \cite{madhusudana2022image} & 0.532 & 0.678 & 0.732 & \textbf{0.073} & \textbf{0.052} & 855 \\
        ARNIQA \cite{agnolucci2024arniqa} & 0.544 & 0.694 & 0.739 & \underline{0.074} & \textbf{0.052} & 855 \\
        CLIP-IQA+ \cite{wang2023exploring} & \underline{0.551} & \underline{0.709} & \underline{0.747} & 0.111 & 0.089 & 895 \\
        QualiCLIP \cite{agnolucci2024qualityaware} & \textbf{0.570} & \textbf{0.725} & \textbf{0.770} & 0.083 & 0.066 & 901 \\ \bottomrule
    \end{tabular}

    \label{tab:test_model_performance}
\end{table}



\section{Conclusion}

We introduced a novel NR-IQA dataset that offers several significant advantages. First, it is the largest UHD and highest-quality IQA database available. Second, it features annotations at higher resolutions than existing IQA datasets for training NR-IQA models. Third, it offers a unique opportunity to evaluate the performance of IQA methods in challenging and practical conditions.
Our dataset offers a way forward for model development, extending observations from previous works regarding cross-resolution generalization \cite{wiedemann2023KonxCrossresolutionImageb} and high-resolution IQA \cite{huang2024HighResolutionImagec}. Including perceptual quality ratings from expert raters and rich metadata ensures it is comprehensive and reliable for training.
Current computer vision methods are increasingly required to perform efficiently and accurately at high resolutions. We believe NR-IQA should also tackle this challenge, which will significantly advance the field and foster the development of practical NR-IQA models that apply to modern, high-quality photos.

\section*{Acknowledgements}
The dataset creation was funded by the Deutsche Forschungsgemeinschaft (DFG, German Research Foundation) – Project-ID 251654672 – TRR 161.

%
%
\bibliographystyle{splncs04}
\bibliography{main} 

\begin{thebibliography}{10}
\providecommand{\url}[1]{\texttt{#1}}
\providecommand{\urlprefix}{URL }
\providecommand{\doi}[1]{https://doi.org/#1}

\bibitem{agnolucci2024qualityaware}
Agnolucci, L., Galteri, L., Bertini, M.: {Quality-Aware Image-Text Alignment
  for Real-World Image Quality Assessment}. arXiv preprint arXiv:2403.11176
  (2024)

\bibitem{agnolucci2024arniqa}
Agnolucci, L., Galteri, L., Bertini, M., Del~Bimbo, A.: {ARNIQA: Learning
  Distortion Manifold for Image Quality Assessment}. In: Proceedings of the
  IEEE/CVF Winter Conference on Applications of Computer Vision. pp. 189--198
  (2024)

\bibitem{fang2020perceptual}
Fang, Y., Zhu, H., Zeng, Y., Ma, K., Wang, Z.: {Perceptual Quality Assessment
  of Smartphone Photography}. In: Proceedings of the IEEE/CVF Conference on
  Computer Vision and Pattern Recognition. pp. 3677--3686 (2020)

\bibitem{ghadiyaram2015massive}
Ghadiyaram, D., Bovik, A.C.: {Massive Online Crowdsourced Study of Subjective
  and Objective Picture Quality}. IEEE Transactions on Image Processing
  \textbf{25}(1),  372--387 (2015)

\bibitem{goring2023QualityAssessmentHighera}
G{\"o}ring, S., Rao, R.R.R., Raake, A.: {Quality Assessment of Higher
  Resolution Images and Videos with Remote Testing}. Quality and User
  Experience  \textbf{8}(1), ~2 (Dec 2023). \doi{10.1007/s41233-023-00055-6}

\bibitem{hammou2024effect}
Hammou, D., Krasula, L., Bampis, C., Li, Z., Mantiuk, R.: {The Effect of
  Viewing Distance and Display Peak Luminance - HDR AV1 Video Streaming Quality
  Dataset}. In: Proceedings of the International Conference on Quality of
  Multimedia Experience (QoMEX) (2024)

\bibitem{hosu2019effective}
Hosu, V., Goldlucke, B., Saupe, D.: {Effective Aesthetics Prediction with
  Multi-Level Spatially Pooled Features}. In: Proceedings of the IEEE/CVF
  Conference on Computer Vision and Pattern Recognition (CVPR). pp. 9375--9383
  (2019)

\bibitem{hosu2018expertise}
Hosu, V., Lin, H., Saupe, D.: Expertise screening in crowdsourcing image
  quality. In: International Conference on Quality of Multimedia Experience
  (QoMEX). pp.~1--6. IEEE (2018)

\bibitem{hosu2020koniq}
Hosu, V., Lin, H., Sziranyi, T., Saupe, D.: {KonIQ-10k: An Ecologically Valid
  Database for Deep Learning of Blind Image Quality assessment}. IEEE
  Transactions on Image Processing  \textbf{29},  4041--4056 (2020)

\bibitem{huang2024HighResolutionImagec}
Huang, H., Wan, Q., Korhonen, J.: {High Resolution Image Quality Database}. In:
  IEEE International Conference on Acoustics, Speech and Signal Processing
  (ICASSP). pp. 3105--3109 (2024). \doi{10.1109/ICASSP48485.2024.10446520}

\bibitem{RecommendationITUT913}
ITUT: {Recommendation ITUT P.913 (06/2021) Methods for the Subjective
  Assessment of Video Quality, Audio Quality and Audiovisual Quality of
  Internet Video and Distribution Quality Television in Any Environment} (2021)

\bibitem{larson2010most}
Larson, E.C., Chandler, D.M.: {Most Apparent Distortion: Full-Reference Image
  Quality Assessment and the Role of Strategy}. Journal of Electronic Imaging
  \textbf{19}(1),  011006--011006 (2010)

\bibitem{li2020norm}
Li, D., Jiang, T., Jiang, M.: {Norm-in-Norm Loss with Faster Convergence and
  Better Performance for Image Quality Assessment}. In: Proceedings of the ACM
  International Conference on Multimedia. pp. 789--797 (2020)

\bibitem{li2023LSDIRLargeScale}
Li, Y., Zhang, K., Liang, J., Cao, J., Liu, C., Gong, R., Zhang, Y., Tang, H.,
  Liu, Y., Demandolx, D., Ranjan, R., Timofte, R., Van~Gool, L.: {{LSDIR}}: {{A
  Large Scale Dataset}} for {{Image Restoration}}. In: IEEE/CVF Computer Vision
  and Pattern Recognition Workshops (CVPRW). pp. 1775--1787. IEEE, Vancouver,
  BC, Canada (2023). \doi{10.1109/CVPRW59228.2023.00178}

\bibitem{lin2022large}
Lin, H., Chen, G., Jenadeleh, M., Hosu, V., Reips, U.D., Hamzaoui, R., Saupe,
  D.: {Large-Scale Crowdsourced Subjective Assessment of Picturewise Just
  Noticeable Difference}. IEEE Transactions on Circuits and Systems for Video
  Technology  \textbf{32}(9),  5859--5873 (2022)

\bibitem{lin2019kadid}
Lin, H., Hosu, V., Saupe, D.: {KADID-10k: A Large-Scale Artificially Distorted
  IQA Database}. In: International Conference on Quality of Multimedia
  Experience (QoMEX). pp.~1--3. IEEE (2019)

\bibitem{liu2020comprehensive}
Liu, J., Liu, D., Yang, W., Xia, S., Zhang, X., Dai, Y.: {A Comprehensive
  Benchmark for Single Image Compression Artifact Reduction}. IEEE Transactions
  on Image Processing  \textbf{29},  7845--7860 (2020)

\bibitem{madhusudana2022image}
Madhusudana, P.C., Birkbeck, N., Wang, Y., Adsumilli, B., Bovik, A.C.: {Image
  Quality Assessment Using Contrastive Learning}. IEEE Transactions on Image
  Processing  \textbf{31},  4149--4161 (2022)

\bibitem{men2021subjective}
Men, H., Lin, H., Jenadeleh, M., Saupe, D.: {Subjective Image Quality
  Assessment with Boosted Triplet Comparisons}. IEEE Access  \textbf{9},
  138939--138975 (2021)

\bibitem{murray2012ava}
Murray, N., Marchesotti, L., Perronnin, F.: {AVA: A Large-Scale Database for
  Aesthetic Visual Analysis}. In: Proceedings of the IEEE/CVF Conference on
  Computer Vision and Pattern Recognition. pp. 2408--2415. IEEE (2012)

\bibitem{ponomarenko2015image}
Ponomarenko, N., Jin, L., Ieremeiev, O., Lukin, V., Egiazarian, K., Astola, J.,
  Vozel, B., Chehdi, K., Carli, M., Battisti, F., et~al.: {Image Database
  TID2013: Peculiarities, Results and Perspectives}. Signal Processing: Image
  Communication  \textbf{30},  57--77 (2015)

\bibitem{ponomarenko2009tid2008}
Ponomarenko, N., Lukin, V., Zelensky, A., Egiazarian, K., Carli, M., Battisti,
  F.: {TID2008: A Database for Evaluation of Full-Reference Visual Quality
  Assessment Metrics}. Advances of Modern Radioelectronics  \textbf{10}(4),
  30--45 (2009)

\bibitem{radford2021learning}
Radford, A., Kim, J.W., Hallacy, C., Ramesh, A., Goh, G., Agarwal, S., Sastry,
  G., Askell, A., Mishkin, P., Clark, J., et~al.: {Learning Transferable Visual
  Models from Natural Language Supervision}. In: International Conference on
  Machine Learning. pp. 8748--8763. PMLR (2021)

\bibitem{sheikh2005live}
Sheikh, H.R., Wang, Z., Cormack, L., Bovik, A.C.: {LIVE Image Quality
  Assessment Database Release 2}.
  \url{http://live.ece.utexas.edu/research/quality} (2005)

\bibitem{su2020blindly}
Su, S., Yan, Q., Zhu, Y., Zhang, C., Ge, X., Sun, J., Zhang, Y.: {Blindly
  Assess Image Quality in the Wild Guided by a Self-Adaptive Hyper Network}.
  In: Proceedings of the IEEE/CVF Conference on Computer Vision and Pattern
  Recognition (CVPR). pp. 3667--3676 (2020)

\bibitem{sun2017mdid}
Sun, W., Zhou, F., Liao, Q.: {MDID: A Multiply Distorted Image Database for
  Image Quality Assessment}. Pattern Recognition  \textbf{61},  153--168 (2017)

\bibitem{tan2019efficientnet}
Tan, M., Le, Q.: {Efficientnet: Rethinking Model Scaling for Convolutional
  Neural Networks}. In: International Conference on Machine Learning (ICML).
  pp. 6105--6114. PMLR (2019)

\bibitem{virtanen_cid2013_2015}
Virtanen, T., Nuutinen, M., Vaahteranoksa, M., Oittinen, P., Häkkinen, J.:
  {CID2013: A Database for Evaluating No-Reference Image Quality Assessment
  Algorithms} (1),  390--402 (2015). \doi{10.1109/TIP.2014.2378061}

\bibitem{wang2023exploring}
Wang, J., Chan, K.C., Loy, C.C.: {Exploring CLIP for Assessing the Look and
  Feel of Images}. In: Proceedings of the AAAI Conference on Artificial
  Intelligence. vol.~37, pp. 2555--2563 (2023)

\bibitem{wiedemann2018disregarding}
Wiedemann, O., Hosu, V., Lin, H., Saupe, D.: {Disregarding the Big Picture:
  Towards Local Image Quality Assessment}. In: International Conference on
  Quality of Multimedia Experience (QoMEX). pp.~1--6. IEEE (2018)

\bibitem{wiedemann2023KonxCrossresolutionImageb}
Wiedemann, O., Hosu, V., Su, S., Saupe, D.: {KonX: Cross-Resolution Image
  Quality Assessment}. Quality and User Experience  \textbf{8}(1), ~8 (Dec
  2023). \doi{10.1007/s41233-023-00061-8}

\bibitem{ying2020patches}
Ying, Z., Niu, H., Gupta, P., Mahajan, D., Ghadiyaram, D., Bovik, A.: {From
  Patches to Pictures (PaQ-2-PiQ): Mapping the Perceptual Space of Picture
  Quality}. In: Proceedings of the IEEE/CVF Computer Vision and Pattern
  Recognition. pp. 3575--3585 (2020)

\bibitem{zhang2021benchmarking}
Zhang, K., Li, D., Luo, W., Ren, W., Stenger, B., Liu, W., Li, H., Yang, M.H.:
  {Benchmarking Ultra-High-Definition Image Super-Resolution}. In: Proceedings
  of the IEEE/CVF International Conference on Computer Vision. pp. 14769--14778
  (2021)

\bibitem{zhang2023blind}
Zhang, W., Zhai, G., Wei, Y., Yang, X., Ma, K.: {Blind Image Quality Assessment
  via Vision-Language Correspondence: A Multitask Learning Perspective}. In:
  Proceedings of the IEEE/CVF Conference on Computer Vision and Pattern
  Recognition. pp. 14071--14081 (2023)

\end{thebibliography}

\end{document}